\documentclass[final,5p,times,twocolumn,sort&compress]{elsarticle}

\usepackage{setspace}
\usepackage{algorithm}
\usepackage{algorithmicx}
\usepackage{algpseudocode}
\usepackage{amsmath}
\usepackage{amsthm}
\usepackage{framed}
\usepackage{xspace}
\usepackage{warpcol}

\usepackage{graphicx}
\usepackage{epsfig}
\usepackage{subfigure}
\usepackage{dcolumn}
\usepackage{float}
\usepackage[normalem]{ulem}
\useunder{\uline}{\ul}{}
\usepackage{multirow}
\usepackage{listings}
\usepackage{booktabs}

\usepackage{mathtools, amssymb}

\makeatletter 
\renewcommand{\@thesubfigure}{\hskip\subfiglabelskip}
 \makeatother

\begin{document}

\begin{frontmatter}



\title{A More Compact Object Detector Head Network with Feature Enhancement and Relational Reasoning}

\author[label1]{Wenchao Zhang}
\author{Chong Fu\fnref{label1,label2,label3} \corref{cor1}}
\ead{fuchong@mail.neu.edu.cn}
\cortext[cor1]{Corresponding author.}
\author[label1]{Xiangshi Chang}
\author[label1]{Tengfei Zhao}
\author[label4]{Xiang Li}
\author[label5]{Chiu-Wing Sham}

\address[label1]{School of Computer Science and Engineering, Northeastern University, 
  Shenyang 110819, China}
\address[label2]{Engineering Research Center of Security Technology of Complex Network System, Ministry of Education, China}
\address[label3]{Key Laboratory of Intelligent Computing in Medical Image, Ministry of Education, Northeastern University, Shenyang 110819, China}
\address[label4]{Baidu Smart Life Group, NO. 10 Xibeiwang East Street, Ke Ji Yuan, Haidian District, Beijing 100193 China}
\address[label5]{School of Computer Science, The University of Auckland, New Zealand}
\begin{abstract}
Modeling implicit feature interaction patterns is of significant importance to object detection tasks. However, in the two-stage detectors, due to the excessive use of hand-crafted components, it is very difficult to reason about the implicit relationship of the instance features. To tackle this problem, we analyze three different levels of feature interaction relationships, namely, the dependency relationship between the cropped local features and global features, the feature autocorrelation within the instance, and the cross-correlation relationship between the instances. To this end, we propose a more \textbf{c}ompact \textbf{o}bject \textbf{d}etector \textbf{h}ead network (CODH), which can not only preserve global context information and condense the information density, but also allows instance-wise feature enhancement and relational reasoning in a larger matrix space. Without bells and whistles, our method can effectively improve the detection performance while significantly reducing the parameters of the model, \emph{e.g.}, with our method, the parameters of the head network is $0.6\times$ smaller than the state-of-the-art Cascade R-CNN, yet the performance boost is 1.3\% on COCO test-dev. Without losing generality, we can also build a more lighter head network for other multi-stage detectors by assembling our method.

\end{abstract}




\end{frontmatter}


\section{Introduction}

\begin{figure}[h]
\centering
\includegraphics[width=0.450\textwidth]{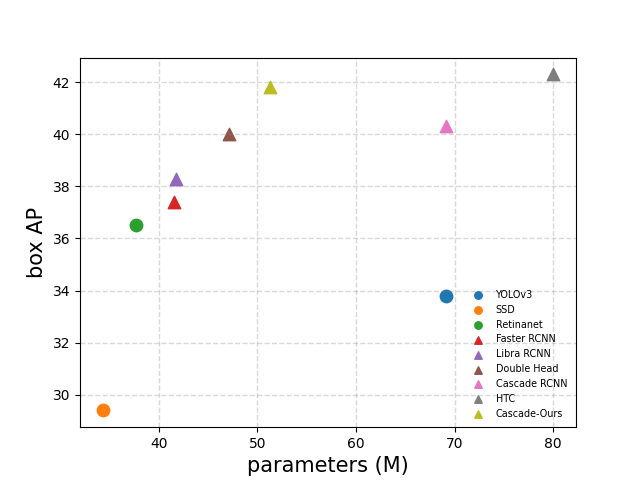}
   \caption{Accuracy-complexity comparison of different detectors on COCO validation dataset. In this figure, the circular mark  represents the one-stage detector, and the triangle mark represents the multi-stage detector. For the multi-stage detector, we use ResNet-50 as the backbone network.}
\label{fig:short}
\label{fig:fig1}
\end{figure}

In recent years, with the the rapid development of deep learning technology, we have witnessed the prosperity of the computer vision community. Especially the emergence of the groundbreaking AlexNet ~\cite{krizhevsky2017imagenet} brings the development of computer vision into a new era. Subsequently, many excellent works have emerged in a broad range of tasks, \emph{e.g.}, classification, object detection, and segmentation. Among them, object detection has been extensively studied since it is a preparatory task for many vision applications. Generally, the prevailing object detection models can be categorized into multi-stage and one-stage methods. The one-stage method is simple and fast, which directly performs dense detection on the features captured by the backbone network. Alternatively, the multi-stage object detection architecture can obtain better accuracy by fine-tuning the bounding box candidates twice, but it also makes the network more complicated with using more hand-crafted components and has more parameters (as shown in Fig.~\ref{fig:fig1}). In contrast, there is an apparent distinction between the two methods, that is, the head network of the two-stage method lacks feature interaction between instances since it performs region-wise detection.

\begin{figure*}[!htb]
  \centering
    \subfigure[Pipeline of one-stage method, (\emph{i.e.}, dense detection)]{\includegraphics[width=0.35\textwidth]{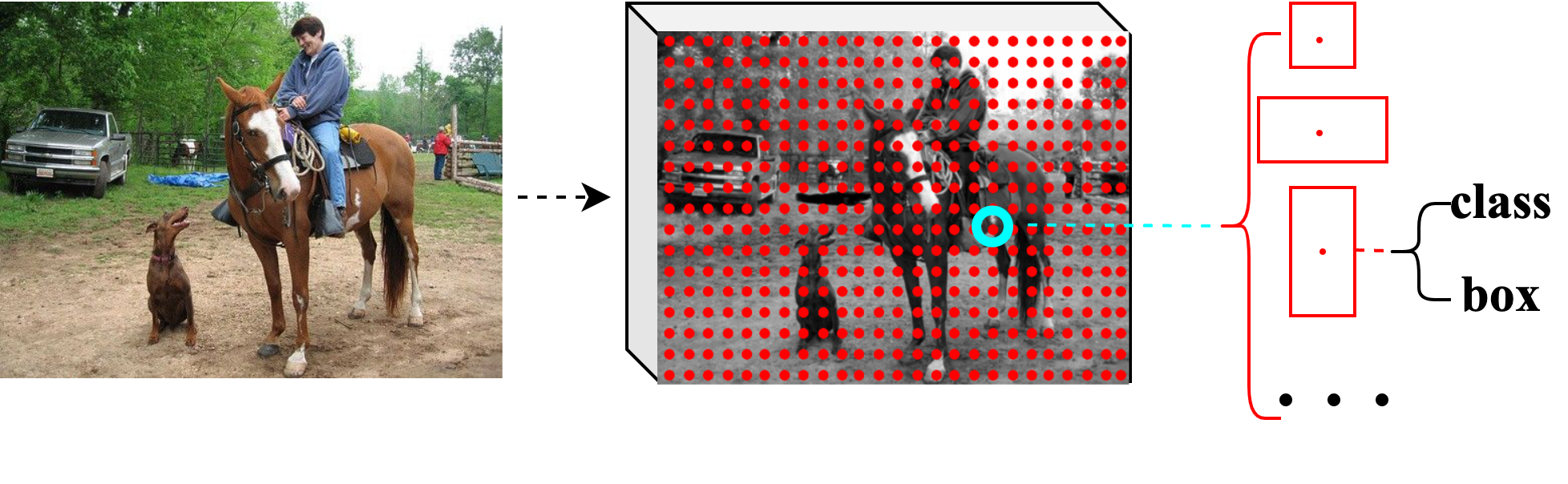}}\hspace{6mm} 
    \subfigure[Pipeline of two-stage method, (\emph{i.e.}, instance-wise detection)]{\includegraphics[width=0.58\textwidth]{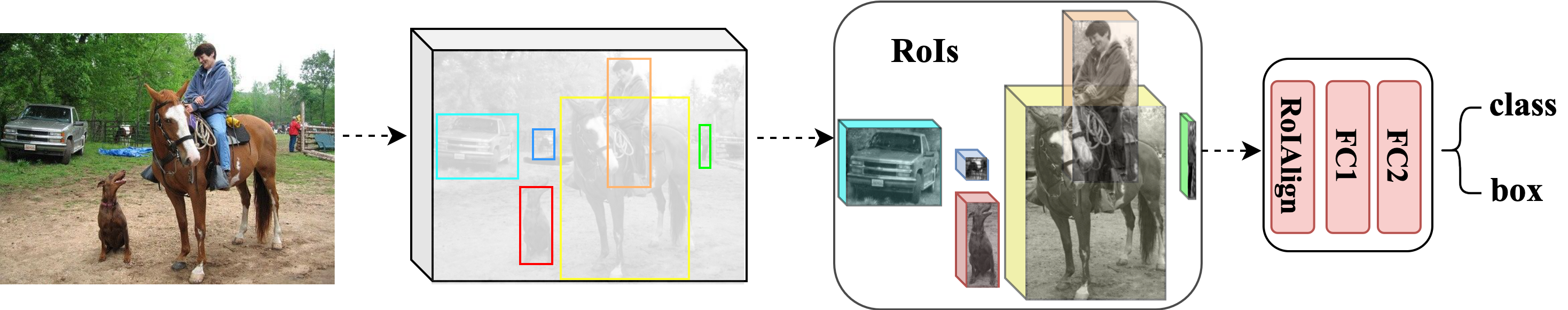}} \\
    \subfigure[Pipeline of ours]{\includegraphics[width=0.7\textwidth]{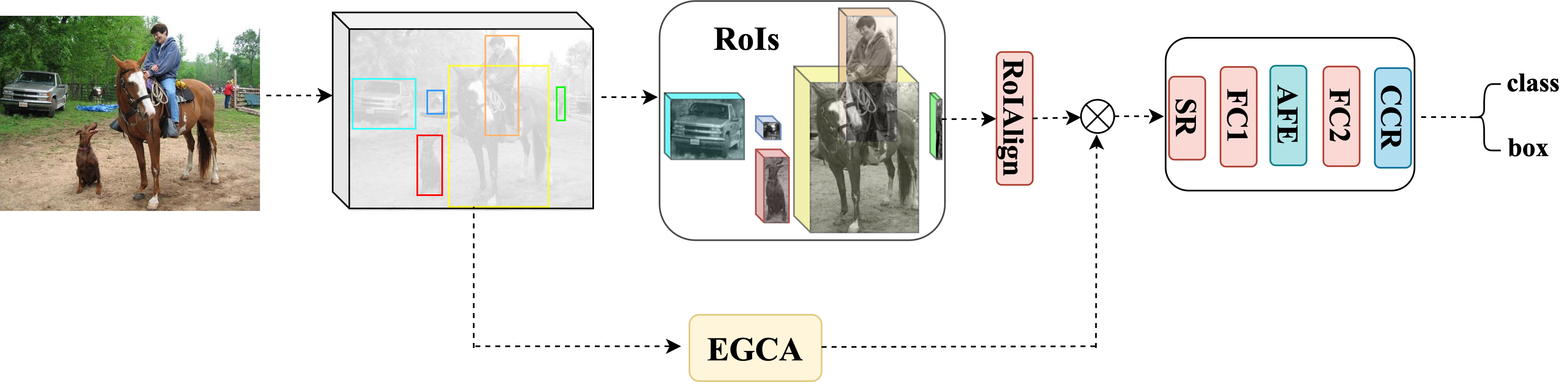}} 
    \caption{Comparisons of different object detection pipelines. (a) In dense detectors, they predict a set of anchor boxes in each grid(\emph{e.g.}, YOLO ~\cite{redmon2016you}) or each pixel(\emph{e.g.}, SSD ~\cite{liu2016ssd}). (b) In instance-wise detectors, they first crop the region of interest from the global feature map, and then perform region-wise detection(\emph{e.g.}, FPN ~\cite{lin2017feature}). (c) Our proposed CODH enhances the interaction of instance features by assembling our proposed EGCA, SR, AFE and CCR. And $\otimes$ stands for channel-wise multiplication. }
  \label{fig:fig2}
\end{figure*}

For clarity, we first briefly review the pipelines of one-stage method (as shown in Fig.~\ref{fig:fig2} (a)) two-stage method (as shown in Fig.~\ref{fig:fig2} (b)). We can simply divide the two-stage method into three parts: global feature extraction (\emph{i.e.}, backbone network), RPN (Region Proposal Network) and subsequent instance-wise classification and localization (\emph{i.e.}, RoI (Region of Interest) head). From Fig.~\ref{fig:fig2} (b), we can see that the two-stage method processes each instance separately after the RoI is cropped from the global feature, which will bring about the head network lacking two types of information interaction: \emph{(1) Interaction between cropped features of local receptive fields and global background information. (2) Interaction between different instances. Intuitively, we believe that these two kinds of interactive information will promote the detection ability of the network. Simultaneously, we empirically prove that in the RoI head of the two-stage method, the representation ability of features can be enhanced by learning the relationship between adjacent features within the instances. Analogously, we  conclude that (3) there is still a lack of interaction of autocorrelated information.} 

To mitigate the drawbacks mentioned above, in this paper, we propose a novel and efficient lightweight object detection head (as shown in Fig.~\ref{fig:fig2}(c)) for two-stage object detectors. Specifically, in terms of inference (1), we propose the Enhanced Global Context Awareness (EGCA) module. It first uses the lightweight attention mechanism to refine the global features captured by the backbone network and then fuses these global features to calibrate the cropped RoI features. Besides, we propose the Autocorrelation Feature Enhancement (AFE) module to handle the inference (2). It first performs dimensional transformation on the 1D features of the instance for learning the correlation of adjacent features through convolution operations in a multi-dimensional space, which can not only take advantage of the parameter sharing of convolution operation, but also facilitate our learning of adjacent features. Then we use a pattern similar to the Inverted Residuals~\cite{sandler2018mobilenetv2} architecture to embed the instance features into a larger metric space for autocorrelation feature enhancement. Naturally, how to establish the interactive information between instances is an orthogonal problem of inference (2). Thus, we propose a Cross-Correlation Reasoning (CCR) module, which can use the AFE module to learn the interactive information between instances by simply transposing the input instance features. Simultaneously, to overcome the paradox of performance and complexity trade-off, we propose the Spatial Reduction (SR) module and Channel Reduction (CR) module to further compress the features after RoIAlign. The proposed CODH is general. We apply it to various prevalent object detection frameworks, including Faster R-CNN~\cite{ren2015faster}, Libra R-CNN~\cite{pang2019libra}, Double Head R-CNN~\cite{wu2020rethinking} and Cascade R-CNN~\cite{cai2018cascade}. Without bells and whistles, by assembling the four lightweight modules we proposed, the accuracy of the two-stage method can be effectively improved with a lighter head network, \emph{e.g.}, as shown in Fig.~\ref{fig:fig1}, by installing our head network, Cascade R-CNN can obtain comparable results with HTC~\cite{chen2019hybrid}, but has much smaller parameters. 

In summary, the main contributions of this work are highlighted as follows:

1. Our study reveals that in the head network of the two-stage method exists three kinds of feature interaction information which needs to be strengthened, namely, the interaction between instance and global feature, and feature interaction within instance and between instances.

2. We propose the CODH method to alleviate the above problems by introducing global context information and learning implicit relationship patterns between adjacent features of instances.

3. The approach we proposed is very lightweight and plug-and-play, which can effectively reduce the amount of parameters and calculations while bringing clear performance gain.

The rest of this paper is organized as follows. In Section ~\ref{sec:sec2}, we briefly review related work on object detection, feature enhancement and relational reasoning. In Section ~\ref{sec:sec3}, we introduce our method in detail from EGCA, SR and CR, AFE, and CCR. In Section ~\ref{sec:sec4}, numerous experiments and analysis of the results are elaborated. Finally, we conclude this paper in Section ~\ref{sec:sec5}.

\section{Related Work}
\label{sec:sec2}
In this section, we briefly sort out the relevant literature from three parts: object detection, feature enhancement, and relational reasoning, since we improve the performance of the object detector by enhancing the internal features of the instances and reasoning about the feature relationships between the instances.

\subsection{Object Detection}

The goal of object detection is to separate the foreground from the background in the image, and to further classify and locate different instance objects simultaneously. The one-stage object detector avoids clipping the region of interest, and performs dense detection on the entire feature map. For example, YOLO~\cite{redmon2016you} divides the global features of different stages into grids of different sizes, and then predicts the objects contained in each grid respectively. SSD~\cite{liu2016ssd} assigns several anchor boxes of different sizes and scales to each position on the feature map, and uses the feature maps of different stages in the network to classify and regress each anchor box. More recently, FCOS~\cite{tian2019fcos} removes the pre-defined anchor box, and uses the idea of semantic segmentation to solve the detection problem pixel by pixel. By contrast, the two-stage method first extracts RoI features, and then performs instance-wise classification and position regression. For instance, Faster R-CNN~\cite{ren2015faster} needs to generate category-agnostic region proposals through RPN, and then use a heavy head network to assign specific categories to each proposal and fine-tune its positioning coordinates. Furthermore, numerous efforts have since continued to push the boundaries of two-stage methods. Some recent frameworks ~\cite{he2017mask,liu2018path,cai2018cascade,huang2019mask,chen2019hybrid,cheng2020boundary,zhang2020mask,qiao2020detectors} add new branches to Faster R-CNN head network for multi-task joint training. Moreover, some studies ~\cite{lin2017feature,li2017light,hu2018relation,pang2019libra,wang2019region,relationnetplusplus2020} attempt to solve the misalignments caused by the original design of the two-stage object detectors that contain many hand-crafted components. In summary, the fully convolutional proposal-free network pipeline of the one-stage method can learn the interaction of different instances to enhance the detection accuracy yet its accuracy is still low since it does not filter out those messy backgrounds. However, the region-wise detection process of the two-stage method weakens the ability to learn these dependency information. Therefore, we hope to improve the detection performance by strengthening the learning of this relationship information in the two-stage method.

\subsection{Feature Enhancement}

Feature enhancement has been proven to be an effective method for learning hidden patterns within features. Some approaches ~\cite{he2016deep,xie2017aggregated,gao2019res2net,zhang2020resnest} utilize identity mapping to avoid the problem that the model accuracy will slowly reach saturation and then quickly degenerate during the training process of convolutional neural networks as the network depth increases. More recently, for application scenarios with limited computing resources, some lightweight networks ~\cite{howard2017mobilenets,sandler2018mobilenetv2,zhang2018shufflenet,ma2018shufflenet,tan2019efficientnet,han2020ghostnet} utilize efficient convolution operations such as depthwise separable convolution or group convolution to establish efficient feature enhancement methods. While in ~\cite{hu2018squeeze,li2019selective,wang2020eca}, they attempt suppressing invalid features and enhancing important features by learning the feature interaction in spatial or channel. In contrast, non-local perception methods ~\cite{dai2017deformable,wang2018non,zhu2019asymmetric} model the dependency between long-distance features through unrestricted feature perception methods. However, the above methods are mostly used in classification and semantic segmentation tasks, and we empirically proved that in the head network of the two-stage method, the convolution operation can also be used to enhance the interaction of instance features through dimensional transformation.

\subsection{Relational Reasoning}

In the literature, learning the interactive information between different instances has demonstrated to offer great potential in enhancing the predictive ability of deep networks. In ~\cite{santoro2017simple}, a Relation Networks (RNs) was proposed for relational reasoning, which utilized a learnable neural network module to find potential relationships between any pairwise objects. Relation Distillation Networks ~\cite{deng2019relation} tried to learn the object interaction relationship in the spatio-temporal context of the video stream for enhancing the feature of the object proposal in reference frames. Graph Relation Network ~\cite{kang2020graph} attempted to explore the neighborhood semantic relationships between learning samples, and then utilized this relationship to embed samples into different metrics space for clustering. 
Analogously, the relation network ~\cite{hu2018relation}, inspired by the transfromer architecture ~\cite{vaswani2017attention}, uses the multi-head attention method and introducing geometric weight information to perform joint reasoning on all object instances. However, it greatly increases the complexity of the network and requires computation at billions of FLOPs. From the above discussion we can see that it is useful to utilize the implicit relationship between objects whether in the field of natural language processing or computer vision. 
Nonetheless, the exploration of this implicit interaction relationship often requires complex network design, which will also significantly increase the occupation of computing resources. In this regard, we establish a lightweight relational reasoning method to mine the interactive relationship between adjacent features of instance objects to enhance the predictive ability of the network.

\section{Methods}
\label{sec:sec3}
In this section, we will introduce the proposed Efficient Feature Enhancement Head in detail, which contains four sub-modules: Enhanced Global Context Awareness, Spatial Reduction and Channel Reduction, Autocorrelation Feature Enhancement, and Cross-Correlation Reasoning.

\subsection{Enhanced Global Context Awareness}
\label{sec:sec3.1}

\begin{figure}[!htb]
  \centering
    \subfigure[Pipeline of EGCA]{\includegraphics[width=0.45\textwidth]{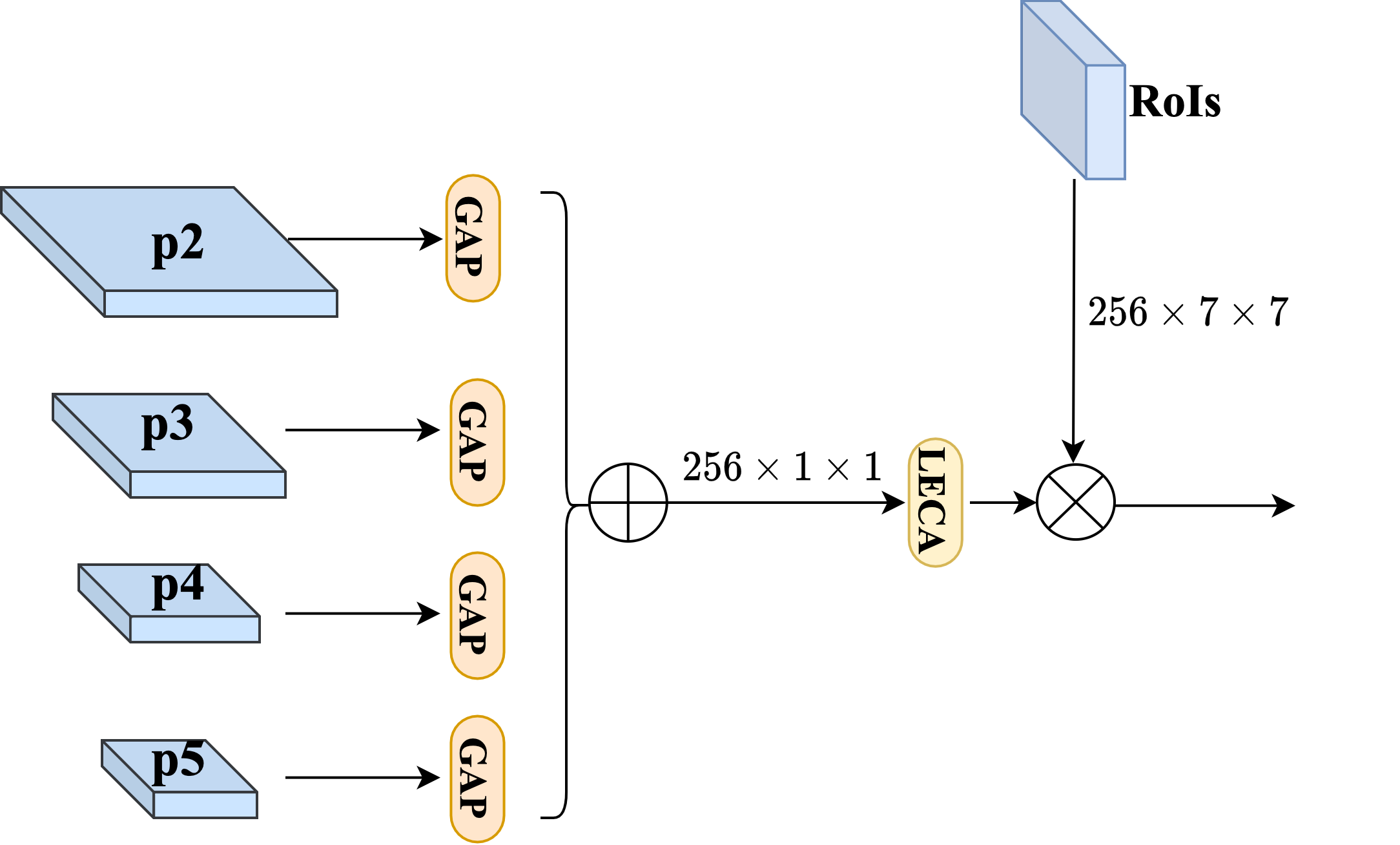}} \\
    \subfigure[An overview of LECA pipeline]{\includegraphics[width=0.48\textwidth]{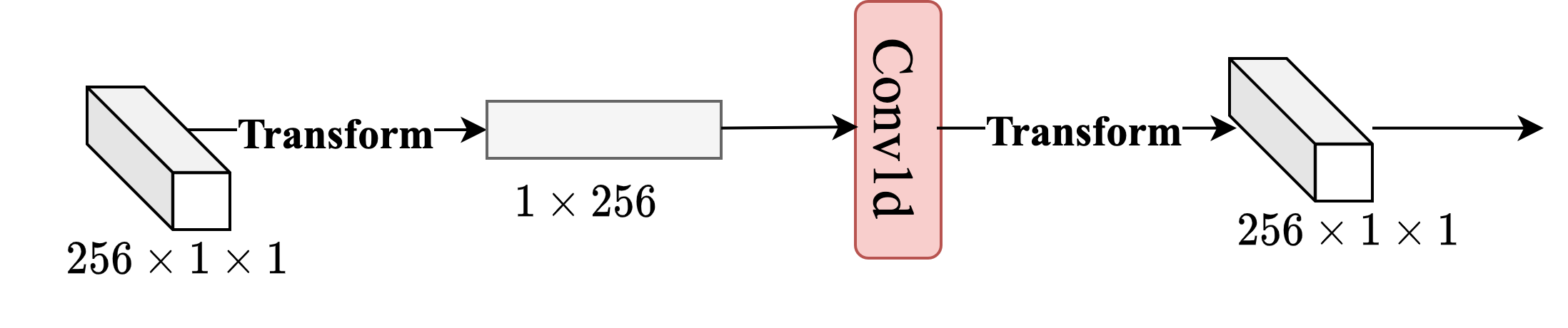}}
    \caption{An overview of EGCA pipeline. (a) In EGCA pipeline, we use the global feature pyramid $\{p2, p3, p4, p5\}$ to refine the global context information respectively, where $\oplus$ stands for element-wise addition and $\otimes$ stands for channel-wise multiplication. (b) In LECA pipeline, we do not need the skip connection structure used in ECA~\cite{wang2020eca} and the feature transformation process is to meet the format of conv1d input.}
  \label{fig:fig3}
\end{figure}

Recently, GCA R-CNN~\cite{zhang2020global} proposes to use dense connections and four parallel global context awareness modules to mitigate the impact of the lack of global information in FPN ~\cite{lin2017feature}, whereas it adds more parameters and calculations. In this paper, inspired by the ECA (Efficient Channel Attention) module~\cite{wang2020eca}, we establish a more lightweight Enhanced Global Context Awareness (EGCA) method. Specifically, as shown in Fig.~\ref{fig:fig3} (a), we first use the feature pyramid $\{p2, p3, p4, p5\}$ to perform preliminary global context condensing through the GAP (Global Average Pooling) layer, which can effectively reduce the feature scale while retaining global information. \emph{It is worth noting that, different from the ECA module, our input feature dimension is 1D, and there is no need to do channel-wise multiplication with the input feature at the end (as shown in Fig.~\ref{fig:fig3} (b)), and to differentiate our approach from ECA, we denote the method used in this paper as LECA (Linear ECA).} Subsequently, the four different levels of global context information are fused by element-wise addition, and then we use one LECA module to further refine the fused features. Finally, the refined global context features are channel-wise multiplied with each RoI feature to perform feature calibration. 

{\bf Parameter count for EGCA: } As mentioned above, we know that in EGCA we only use one LECA module to refine global context information. And in LECA, it uses only one conv1d layer with a kernel size of $k$ to learn the relationship between adjacent features of the input 1D features. Therefore, the parameter quantity of EGCA module is $k$, and inspired by ECA we set $k=5$ in our experiment.

\subsection{Spatial Reduction and Channel Reduction}
In Faster R-CNN head network, after RoIAlign, each RoI is normalized to a uniform size ($256\times7\times7$), and then two FC (fully connected) layers are employed for further encoding. \emph{However, it is worth noting that compared to the second FC layer, the first FC layer has a huge amount of parameters ($256\times7\times7\times1024$ vs. $1024\times1024$), which makes Faster R-CNN have a heavy head. In light of this state of affairs, for the purpose of reducing the amount of parameters of the FC1 layer, we compress the RoI features after RoIAlign in spatial and channel respectively.} Simultaneously, we define the \emph{spatial compression multiplier} $\alpha$ ($\alpha \ge 1$) and the \emph{channel compression multiplier} $\beta$ ($0 \le \beta \le 1$) to control the compression ratio respectively, which means the size of the compressed RoI is $\beta\times256\times\alpha\times\alpha$. 
Specifically, for Spatial Reduction (SR), with the goal of strengthening the feature connection in the channel and reducing the amount of parameters used, we first deform the output feature tensor of RoIAlign to $256\times49$, and then use conv1d with a kernel size of $k^{'}$ and a stride size of $s$ to perform the compression operation. For Channel Reduction (CR), we only use $1\times1$ convolution to perform reduction operations. In our experiment, $k^{'}=5$, $s=2$, $\alpha=5$, $\beta=None$ (which means no channel reduction is performed).

{\bf Parameter count for SR and CR: } From the above description, we can see that we only perform spatial reduction, and thus the newly added parameter amount is $256\times256\times k^{'}$. Continuously, if the channel reduction is performed, the parameter amount will be increased by $256\times\beta\times256$ additionally. Furthermore, after using SR and CR, the parameter amount of the first FC layer can be compressed to $\beta\times\alpha^2\times256\times1024$.

\subsection{Autocorrelation Feature Enhancement}
\label{sec:sec3.3}
\begin{figure}[htbp]
\centering
\includegraphics[width=0.45\textwidth]{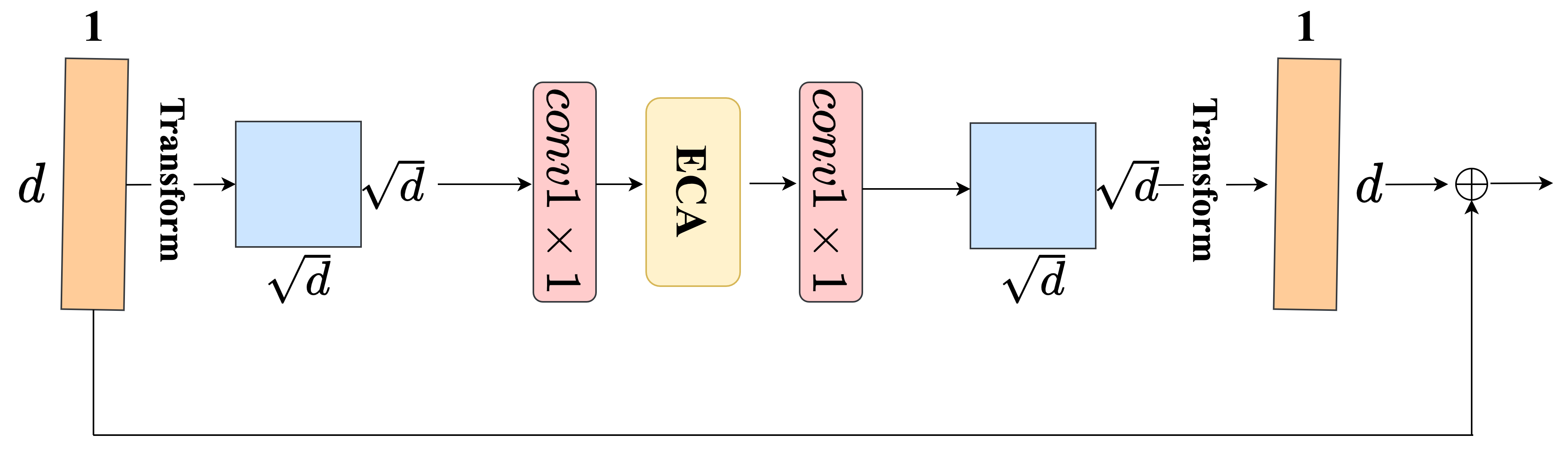}
   \caption{An overview of AFE pipeline. For simplicity, we only use one object instance to illustrate the AFE pipeline. and $\oplus$ stands for element-wise addition.}
\label{fig:short}
\label{fig:fig4}
\end{figure}

As analyzed in the previous section, in Faster R-CNN pipeline, before decoupling the classification and regression positioning tasks, it performs further instance-wise feature encoding by applying two shared FC layers. Actually, this learning process is to enhance the autocorrelation features of each instance. Nonetheless, the autocorrelation features in this pipeline are limited to the 1024-d metric space, which will reduce the learning ability of the neural network. Following that intuition, as shown in Fig.~\ref{fig:fig4}, we introduce a new architectural unit, which we term the Autocorrelation Feature Enhancement (AFE) block, with the goal of expanding the learning space of autocorrelation features. Specifically, we first map instance features to 2D space through dimensional transformation (as shown in Fig.~\ref{fig:fig5} (a)), and then use two $1\times1$ convolution layers to generate richer features to enhance the representative ability of instance features. Note that after the first convolutional layer, we insert an ECA module to perform feature calibration. The feature enhancement process is defined as follows:
\begin{displaymath}{}
\label{eq:eq1}
\mathcal{X}_T=T(\mathcal{X}),\tag{$1$}{}
\end{displaymath}
\begin{displaymath}{}
\label{eq:eq2}
\mathcal{Y}=\mathcal{F}_{2}^{k^{''}}(ECA(\delta(\mathcal{F}_{1}^{k^{''}}(\mathcal{X}_T)))),\tag{$2$}{}
\end{displaymath}
\begin{displaymath}{}
\label{eq:eq3}
\mathcal{Y}_{out}=IT(\mathcal{Y})+\mathcal{X},\tag{$3$}{}
\end{displaymath}
where $\mathcal{X}\in\mathbb{R}^{N \times d}$ in Eq.~(\ref{eq:eq1}) represents the input feature, $N$ and $d$ are the number of RoI, and the input feature dimension. $T(\cdot)$ stands for the dimensional transformation operation, and $\mathcal{X}_T\in\mathbb{R}^{N \times C \times H \times W}$ is the feature after the dimensional transformation, where $C$, $H$ and $W$ are channel dimension, height and width($H=W=\sqrt{d}$). In Eq.~(\ref{eq:eq2}), $\mathcal{F}_{1}$ and $\mathcal{F}_{2}$ are two convolutional layers, and the superscript $k^{''}$ represents the size of the convolution kernel. In addition, $\delta(\cdot)$ stands for the ReLU (Rectified Linear Unit) activation function, and $ECA$ stands for ECA module. Please refer to ~\cite{wang2020eca} for further details. In the AFE module, we use a convolution mode similar to the Inverted Residual Block~\cite{sandler2018mobilenetv2} to excite and squeeze the input features with an expansion rate of $r$, but the difference is that our structure does not contain $3\times3$ depthwise separable convolutional layer, since our focus is on modeling the relationship between neighbors of the 1D input features. For instance, assuming that $\mathcal{X}_{f1}$ is the output of $\mathcal{F}_{1}$, so we use $\mathcal{F}_{2}$ to model the point-wise relationship of $\mathcal{X}_{f1}$. In Eq.~(\ref{eq:eq3}), $IT(\cdot)$ represents the inverse dimensionality transformation, which is used to map the output feature back to 1024-d.

{\bf Parameter count for AFE:} Assuming that the input feature shape of AFE is $(N, d)$, and thus the shape after dimensional transformation is $(N, 1, \sqrt{d}, \sqrt{d})$. Then after the convolutional layer $\mathcal{F}_{1}$ with the expansion rate $r$ and the kernel size $k$, the output feature shape is $(N, r, \sqrt{d}, \sqrt{d})$, so the parameter of $\mathcal{F}_{1}$ is $k^{''} \times k^{''} \times 1 \times r$. Besides, the convolution kernel size of Conv1d in the ECA module is $k^{'''}$. Subsequently, we use $\mathcal{F}_{2}$ to map $(N, r, \sqrt{d}, \sqrt{d})$ back to $(N, 1, \sqrt{d}, \sqrt{d})$, so the parameter of $\mathcal{F}_{2}$ is $k^{''} \times k^{''} \times r \times 1$. In addition, the Eq.~(\ref{eq:eq1}) and Eq.~(\ref{eq:eq3}) are only dimensional transformations and do not require any parameters, so the total parameter quantity of our AFE module is $2 \times k^{''} \times k^{''} \times r + k^{'''}$, and in the experiment we set $k^{''}=1$, $k^{'''}=3$, $r=16$, $d=1024$.

\subsection{Cross-Correlation Reasoning}
\label{sec:sec3.4}
As described in Section~\ref{sec:sec3.3}, the last two shared FC layers in Faster R-CNN are actually enhancements to the internal autocorrelation features of each instance, and does not model the cross-correlation between different instances. In ~\cite{hu2018relation}, for exploring the relationship between instances, the instance features are converted into value, key, and query respectively, and geometric feature is introduced to apply the transformer structure proposed in ~\cite{vaswani2017attention}, therefore sharing the similar spirit of ours. Nevertheless, it adds a large number of parameters, resulting in a heavier RoI head. We deduce that this kind of interactive information can be inferred only by using the features of the object proposal itself. But how to model this interactive information more efficiently is a problem worth exploring. 
\begin{figure}[!htb]
  \centering
    \subfigure[An example of feature transformation in AFE]{\includegraphics[width=0.28\textwidth]{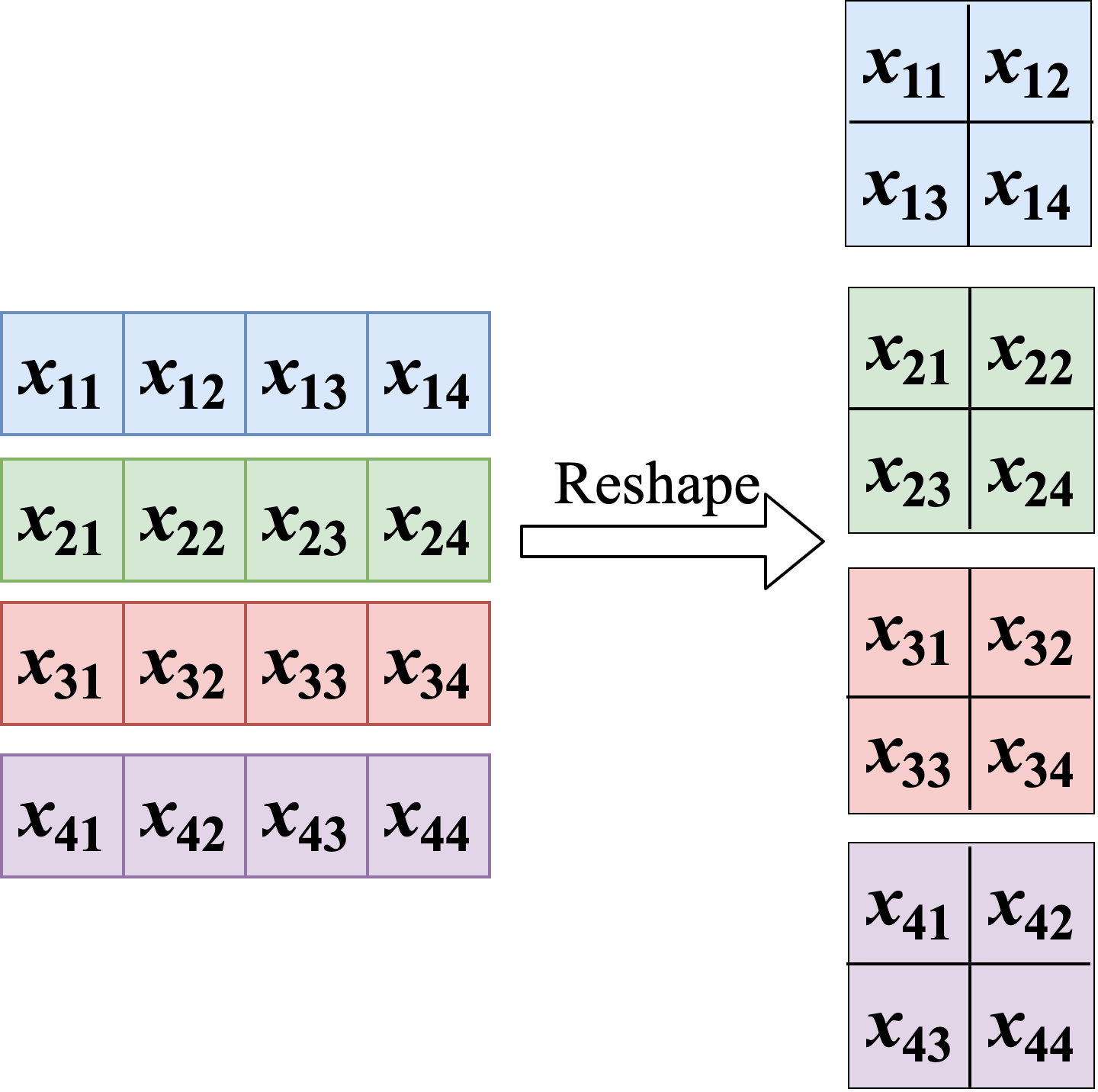}} \\
    \subfigure[An example of feature transformation in CCR]{\includegraphics[width=0.45\textwidth]{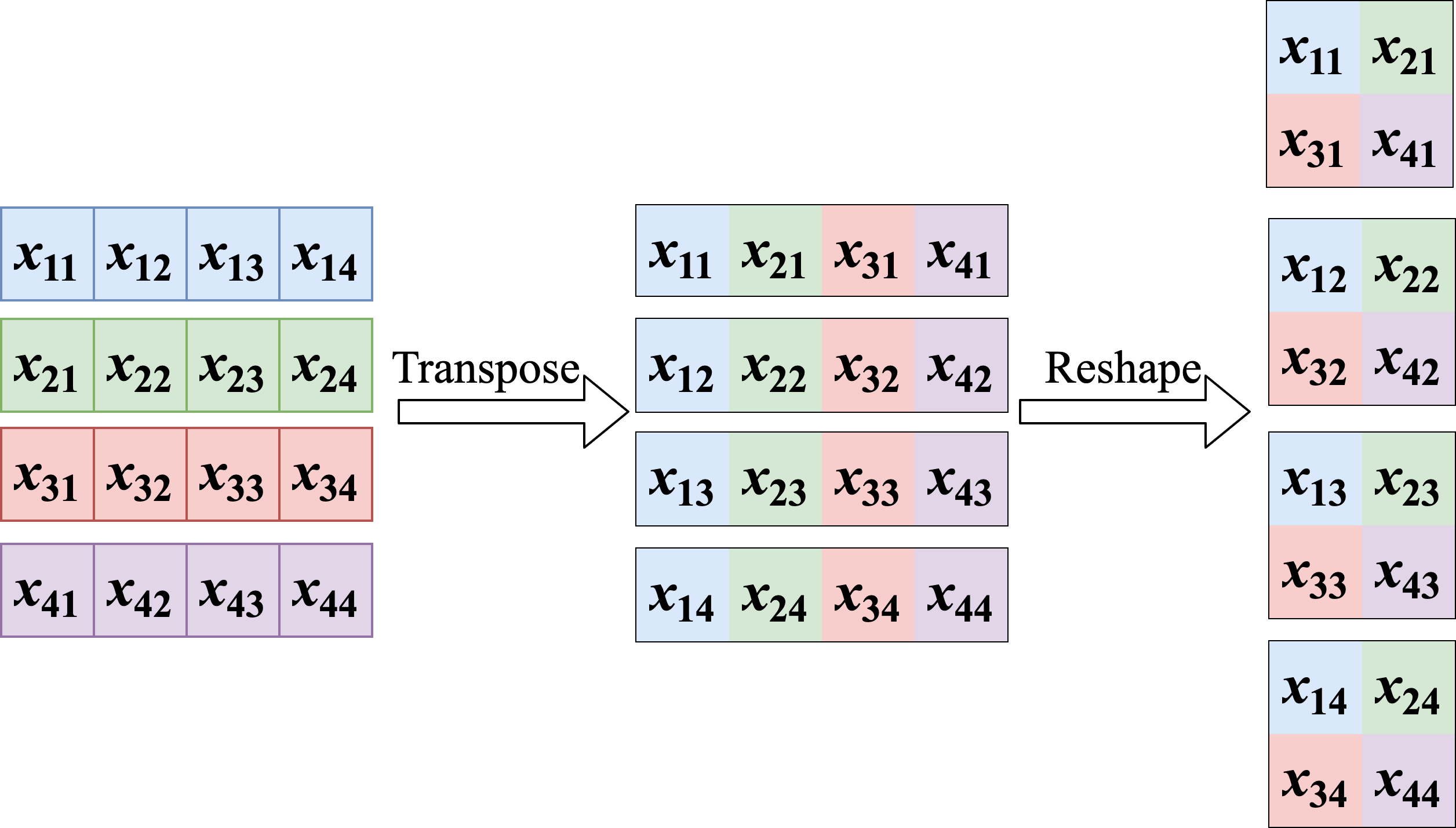}}
    \caption{Examples of feature transformation for AFE and CCR. For simplicity, we use $\mathcal{X}\in\mathbb{R}^{N \times d}, N=d=4$ to describe the feature transformation process. (a) In the AFE module, each feature channel after transformation only contains the features of one instance. (b) In the CCR module, each feature channel after transformation contains adjacent features of all instances.}
  \label{fig:fig5}
\end{figure}

In ~\cite{hu2018relation}, in order to obtain the features of the two different attributes of key and query, it undergoes two FC layers to transform the input features, and learns multiple relation features by grouping the input features in the multi-head attention module, finally, applies a $1\times1$ convolutional layer to map the learned multiple relation features to the same dimensions as the input features. Compared with FC operations, the convolution method can not only greatly reduce the amount of parameters required through parameter sharing, but also allow us to select receptive fields of different scales by adjusting the size of the convolution kernel to determine the adjacent feature range that need to be considered. Hence, like the AFE module, we first transform the object proposal feature obtained after RoIAlign into 2D space, and then use the convolution operation to model the interaction between different instances. In the AFE module, our input is $\mathcal{X}\in\mathbb{R}^{N \times d}$, but in CCR, we want to model the relationship between $\mathcal{X}(m, i)$ and $\mathcal{X}(n, i)$, $((m, n)\in\mathbb{R}^{N}, i\in\mathbb{R}^{d})$. To this end, we transform the input $\mathcal{X}$ to $\mathcal{X}^{T}\in\mathbb{R}^{d \times N}$ by a simple transpose transformation (as shown in Fig.~\ref{fig:fig5} (b)), and thus the interaction between instances can be modeled by point-wise convolution and ECA operation. \emph{Therefore, compared with the AFE module, our CCR module only adds two transpose operations before the input and the final output, which can be defined as follows}:
\begin{displaymath}{}
\label{eq:eq4}
\mathcal{X}=\mathcal{X}^{T},\tag{$4$}{}
\end{displaymath}
\begin{displaymath}{}
\label{eq:eq5}
\mathcal{Y}_{out}=\mathcal{Y}_{out}^{T}.\tag{$5$}{}
\end{displaymath}

{\bf Parameter count for CCR:} Compared with AFE module, we only simply add two transpose operations in the CCR module, and thus the parameter quantity of our CCR module is the same as the AFE module, namely, $2 \times k^{''} \times k^{''} \times r + k^{'''}$.

\section{Experiments}
\label{sec:sec4}
In this section, we first introduce the evaluation criteria and hyperparameter settings of the experiment, and then conduct  extensive comparative experiments to verify our method. In the experimental deployment, we first perform optimization experiments on the hyperparameters and structures of the different proposed modules, and then analyze the contribution of different modules to the model through ablation experiments. Finally, we compare our method with state-of-the-art counterparts on COCO. 

\subsection{Dataset and Metrics}

All our experiments are compared on the large scale detection COCO dataset~\cite{lin2014microsoft} provided by Microsoft. As one of the most commonly used object detection benchmark datasets, it provides $115k$ annotated images for training and $5k$ minival datasets. Simultaneously, it also has $20k$ annotated but unavailable test-dev images, which need to upload the results to the server to verify the performance of the model online. Additionally, we use the standard metric provided by COCO to measure our algorithm, which uses the average AP value under different IoU (Intersection over Union) thresholds as the final result.

\subsection{Implementation Details}

All our experiments are deployed on mmdetection toolkit~\cite{mmdetection}, using NVIDIA titan xp GPU with the mini-batch size of two images. For the ablation experiments, we train detectors for 12 epochs with an initial learning rate of 0.005, and decrease it by 0.1 after 8 and 11 epochs, respectively. Note that unless otherwise specified, we use Faster R-CNN with ResNet50~\cite{he2016deep} as our baseline for comparison, and for other parameter settings, we use the default values in mmdetection. Note that unless otherwise specified, we only use one AFE or CCR module after the second shared FC layer in the Faster R-CNN to perform ablation experiments, and the expansion rate $r$ in the AFE and CCR modules is $16$.

\subsection{Ablation Study}

{\bf Different designs of EGCA:} Since there are four different stages of global features $\{p2, p3, p4, p5\}$, we try two different global feature fusion strategies for the design of the EGCA module: fusion first and extraction first. The fusion first pipeline is described in Section~\ref{sec:sec3.1}. And for extraction first, we first reduce the spatial scale of $\{p2, p3, p4, p5\}$ through the GAP layer, and then use four parallel LECA modules after GAP to further refine the features of $\{p2, p3, p4, p5\}$ respectively. Subsequently, the four different levels of global context information are fused by element-wise addition, and finally they are channel-wise multiplied with local features to perform feature calibration.

From Table~\ref{tab:my-table1}, we can see that both the fusion first and the extraction first methods can significantly improve the performance of object detection, which verifies the correctness of our hypothesis that the RoI head of the two-stage methods lack global context information. Simultaneously, we can see that the result of the extraction first method is slightly higher than fusion first, which is also reasonable, since there are more learnable parameters in the former method due to the fact that we adopt four parallel LECA modules. But in order to better balance accuracy and model complexity, we use the fusion first method as our default setting.

\begin{table}[t]
\caption{Comparison of different designs of EGCA(\%)}
\label{tab:my-table1}
\setlength{\tabcolsep}{3mm}{
\begin{center}
\begin{tabular}{cccc}
\hline
\multicolumn{1}{c}{\bf Method}  &\multicolumn{1}{c}{\bf AP} &\multicolumn{1}{c}{\bf AP$_{50}$} &\multicolumn{1}{c}{\bf AP$_{75}$}  
\\ \hline 
Baseline     & 37.4 & 58.1  & 40.4   \\
Fusion first & 38.0 & 59.2  & 41.0   \\
Extraction first &  \textbf{38.1} &  \textbf{59.3}  &  \textbf{41.3}   \\ 
\hline
\end{tabular}
\end{center}}
\end{table}

{\bf Different kernel sizes of AFE\&CCR:} Different kernel sizes in AFE can enable the convolutional layer to learn the feature relationships of different receptive fields, and hence we analyze the influence of the receptive field on model performance by setting different convolution kernel sizes. Specifically, we set the convolution kernel size $k^{''}$ of $\mathcal{F}_{1}$ and $\mathcal{F}_{2}$ in the Eq.~(\ref{eq:eq1}) and Eq.~(\ref{eq:eq3}) to \{1, 3\} respectively. Analogously, we do the same experiment on the CCR module.

As shown in Table~\ref{tab:my-table2}, our AFE and CCR modules can effectively improve the performance of the detector when $k^{''}=1$, and different kernel size settings will also affect the performance of these two components. Especially for the CCR module, the model performance will degrade sharply when the kernel size is 3. We infer that this is because the CCR module models feature interactions between different instances, and the direct correlation between different instances is weakly correlated. Consequently, when the learnable feature area increases, it will have a adverse impact on the detection accuracy.

\begin{table}[t]

\caption{Comparison of different kernel sizes of AFE\&CCR(\%)}
\label{tab:my-table2}
\setlength{\tabcolsep}{3mm}{
\begin{center}
\begin{tabular}{ccccc}
\hline
\multicolumn{1}{c}{\bf Method} &\multicolumn{1}{c}{\bf $k^{''}$}  &\multicolumn{1}{c}{\bf AP} &\multicolumn{1}{c}{\bf AP$_{50}$} &\multicolumn{1}{c}{\bf AP$_{75}$}  
\\ \hline 
\multirow{2}{*}{AFE} & 1 & \textbf{38.1} & 59.1  & \textbf{41.6}   \\
                     & 3 & 38.0 & \textbf{59.2}  & 41.1   \\
\hline
\multirow{2}{*}{CCR} & 1 &  \textbf{38.1} &  \textbf{59.0}  &  \textbf{41.4}   \\ 
                     & 3 &  28.3 &  47.5  &  29.4   \\
\hline
\end{tabular}
\end{center}}
\end{table}

{\bf Compared with Inverted Residual Block:} Table 3 shows the results of using the Inverted Residual Block structure in the AFE and CCR modules, since the structure we proposed is very similar to it. From the comparison in Table ~\ref{tab:my-table3}, we can see that in the AFE module, integration of either Inverted Residual Block or our method can improve performance of object detection by a clear margin, but in contrast, our method performs slightly better and has fewer parameters. As for the CCR module, as analyzed in the previous section, the performance of the detector will be greatly reduced due to the use of the $3\times3$ depthwise separable convolutional layer in the Inverted Residual Block.
\begin{table}[t]

\caption{Compared with Inverted Residual Block(\%)}
\label{tab:my-table3}
\setlength{\tabcolsep}{2mm}{
\begin{center}
\begin{tabular}{cccc}
\hline
\multicolumn{1}{c}{\bf Method}  &\multicolumn{1}{c}{\bf AP} &\multicolumn{1}{c}{\bf AP$_{0.5}$} &\multicolumn{1}{c}{\bf AP$_{0.75}$}  
\\ \hline 
FPN baseline     & 37.4 & 58.1  & 40.4   \\
\hline
AFE          &  \textbf{38.1} & \textbf{59.1}  & \textbf{41.6}   \\
AFE-Inverted &  38.0 & \textbf{59.1}  & 41.1   \\
\hline
CCR          &  \textbf{38.1} &  \textbf{59.0}  &  \textbf{41.4}   \\  
CCR-Inverted &  29.2 &  49.0  &  30.6   \\ 
\hline
\end{tabular}
\end{center}}
\end{table}
{\bf Different expansion rates of AFE\&CCR:} The expansion rate in AFE\&CCR determines the size of the metric space for learning the interaction of instance features, therefore, in order to explore the influence of different metric spaces on the feature interactive learning ability of the AFE\&CCR, we set the expansion rate $r$ to $\{8, 16, 32\}$ respectively for analysis.

From the comparison in Table~\ref{tab:my-table4}, we can see that as the expansion rate $r$ increases, the performance of the AFE and CCR modules will increase slightly, but they will quickly become saturated. The reason lies in that our input feature space is small, and thus the feature diversity generated by a larger expansion rate is limited. Simultaneously, we can see that the performance of the AFE and CCR modules is optimal when $r=16$, so we use this value as the default value in all subsequent experiments.
\begin{table}[t]

\caption{Comparison of different expansion rates(\%)}
\label{tab:my-table4}
\setlength{\tabcolsep}{2mm}{
\begin{center}
\begin{tabular}{ccccc}
\hline
\multicolumn{1}{c}{\bf Method} &\multicolumn{1}{c}{\bf $r$}  &\multicolumn{1}{c}{\bf AP} &\multicolumn{1}{c}{\bf AP$_{50}$} &\multicolumn{1}{c}{\bf AP$_{75}$}  
\\ \hline 
\multirow{3}{*}{AFE} & 8 & 37.9 & 58.9  & 41.1   \\
                     & 16 & \textbf{38.1} & 59.1  & \textbf{41.6}   \\
                     & 32 & \textbf{38.1} & \textbf{59.3}  & \textbf{41.6}   \\
\hline
\multirow{3}{*}{CCR} & 8 &  38.0 &  \textbf{59.1}  &  \textbf{41.2}   \\ 
                     & 16 &  \textbf{38.1} &  59.0  &  \textbf{41.4}   \\
                     & 32 &  \textbf{38.1} &  59.0  &  41.3   \\
\hline
\end{tabular}
\end{center}}
\end{table}

\begin{table}[t]

\caption{Effect of AFE\&CCR with ECA module(\%)}
\label{tab:my-table5}
\setlength{\tabcolsep}{2mm}{
\begin{center}
\begin{tabular}{cccc}
\hline
\multicolumn{1}{c}{\bf Method}  &\multicolumn{1}{c}{\bf AP} &\multicolumn{1}{c}{\bf AP$_{0.5}$} &\multicolumn{1}{c}{\bf AP$_{0.75}$}  
\\ \hline 
FPN baseline     & 37.4 & 58.1  & 40.4   \\
AFE     & \textbf{38.1} & 59.1  & \textbf{41.6}   \\
AFE+ECA & \textbf{38.1} & \textbf{59.3} & \textbf{41.6}   \\
\hline
CCR     & \textbf{38.1} & \textbf{59.0}  & \textbf{41.4}   \\
CCR+ECA &  38.0 &  58.9  &  41.4   \\ 
\hline
\end{tabular}
\end{center}}
\end{table}

\begin{table}[t]

\caption{Comparison of different AFE and CCR combinations(\%)}
\label{tab:my-table6}
\setlength{\tabcolsep}{2mm}{
\begin{center}
\begin{tabular}{cccc}
\hline
\multicolumn{1}{c}{\bf Method}  &\multicolumn{1}{c}{\bf AP} &\multicolumn{1}{c}{\bf AP$_{50}$} &\multicolumn{1}{c}{\bf AP$_{75}$}  
\\ \hline 
FC2-AFE-CCR     & 37.7 & 58.9  & 41.1   \\
FC2-CCR-AFE & 37.9 & 58.9  & 41.2   \\
AFE-CCR-FC2     & 38.2 & 59.2  & 41.5   \\
CCR-AFE-FC2 &  38.1 &  58.9  &  41.6   \\ 
AFE-FC2-CCR     & \textbf{38.4} & \textbf{59.3}  & 41.4   \\
CCR-FC2-AFE &  38.2 &  59.2  &  41.6   \\ 
AFE-FC2-AFE     & 38.3 & 59.1  & 41.8   \\
CCR-FC2-CCR     & 37.9 & 58.6  & 40.9   \\
FC2-\{AFE,CCR\}     & 37.9 & 59.0  & 41.1   \\
FC2-\{CCR\_cls,AFE\_reg\}     & 38.0 & 59.1  & 41.5   \\
FC2-\{AFE\_cls,CCR\_reg\}     & 37.9 & 58.8  & 41.1   \\
AFE$\times2$-FC2-CCR$\times2$     & \textbf{38.4} & 59.2  & \textbf{41.9}   \\
\{AFE,AFE\}-FC2-\{CCR,CCR\}     & 38.1 & 59.1  & 41.1   \\
\hline
\end{tabular}
\end{center}}
\end{table}

{\bf AFE\&CCR with ECA:} In order to further enhance the ability of the AFE and CCR modules to represent the implicit patterns of the internal features of the instance, we try to add an ECA module after the $\mathcal{F}_{1}$ layer in the AFE and CCR modules to enhance the connection between the channels of the instance features. From the comparison in Table ~\ref{tab:my-table5}, we can see that the performance of the model does not change significantly after adding the ECA module. The reason is the same as the analysis in the previous section, because our feature space is limited and the correlation between channels is relatively close, so using the ECA module to reweight different feature channels has little effect.

\begin{table}[t]
\caption{Comparison of different \{$\alpha$, $\beta$\} combinations(\%) ("None" means no spatial reduction or channel reduction)}
\label{tab:my-table7}
\setlength{\tabcolsep}{2mm}{
\begin{center}
\begin{tabular}{ccccc}
\hline
\multicolumn{1}{c}{\bf \{$\alpha$, $\beta$\}}  &\multicolumn{1}{c}{\bf AP} &\multicolumn{1}{c}{\bf AP$_{50}$} &\multicolumn{1}{c}{\bf AP$_{75}$} &\multicolumn{1}{c}{\bf Params (M)} 
\\ \hline 
Baseline     & 37.4 & 58.1  & 40.4 & 41.53  \\
\{None, 0.5\}     & 38.1 & 59.4  & 41.0 & 35.14  \\
\{None, 0.25\} & 37.6 & 58.9  & 40.8 & 31.91  \\
\{5, None\}     & \textbf{38.4} & 59.7  & \textbf{41.9} & 35.56  \\
\{4, None\} &  38.2 &  \textbf{59.8}  &  41.3  & 33.20 \\ 
\{3, None\}     & 38.2 & 59.7  & 41.6 & 31.37  \\
\{2, None\}     & 37.3 & 59.0  & 40.0 & \textbf{30.06}  \\
\{5, 0.5\} &  38.0 &  59.2  &  41.1  & 32.32 \\ 
\{4, 0.5\}     & 37.9 & 59.1  & 41.1 & 31.14  \\
\{3, 0.5\}     & 37.5 & 58.9  &  40.9 & 30.22  \\
\hline
\end{tabular}
\end{center}}
\end{table}

 \begin{table}[t]
\caption{Comparison of different variant experiments(\%)}
\label{tab:my-table8}
\setlength{\tabcolsep}{2mm}{
\begin{center}
\begin{tabular}{ccccc}
\hline
\multicolumn{1}{c}{\bf Method}  &\multicolumn{1}{c}{\bf AP} &\multicolumn{1}{c}{\bf AP$_{50}$} &\multicolumn{1}{c}{\bf AP$_{75}$} &\multicolumn{1}{c}{\bf Params (M)} 
\\ \hline 
SR     & \textbf{38.4} & \textbf{59.7}  & \textbf{41.9} & 35.56  \\
SR\_Conv3     & 37.9 & 58.5  & 41.3 & 35.83  \\
SR\_Conv3\_Group & 38.0 & 58.7  & 41.7 & \textbf{35.24}  \\
RoIAlign\_$5\times5$  & 37.2 & 58.1  & 40.3 & \textbf{35.24}  \\
\hline
\end{tabular}
\end{center}}
\end{table}

\begin{table*}[h]
\caption{Effects of each component in our method(\%)}
\label{tab:my-table9}
\setlength{\tabcolsep}{2mm}{
\begin{center}
\begin{tabular}{ccccccccccc}
\hline
\multicolumn{1}{c}{\bf EGCA} &\multicolumn{1}{c}{\bf SR} &\multicolumn{1}{c}{\bf AFE} &\multicolumn{1}{c}{\bf CCR} &\multicolumn{1}{c}{\bf AP} &\multicolumn{1}{c}{\bf AP$_{50}$} &\multicolumn{1}{c}{\bf AP$_{75}$} &\multicolumn{1}{c}{\bf AP$_{s}$} &\multicolumn{1}{c}{\bf AP$_{m}$} &\multicolumn{1}{c}{\bf AP$_{l}$} &\multicolumn{1}{c}{\bf Params (M)} 
\\ \hline 
 & & & & 37.4 & 58.1 & 40.4 &21.2 &41.0 &48.1 & 41.53   \\
 & &\checkmark & & 38.1  & 59.1   & 41.6 & 22.7 & 41.3 & 49.6 & 41.53    \\
 & & & \checkmark& 38.1  & 59.0   & 41.4 & 21.8 & 41.4 & 49.4 & 41.53  \\
 \checkmark & & & & 38.0  & 59.2   & 41.0 & 22.3 & 41.6 & 49.1 & 41.53  \\
 & \checkmark & &  & 37.1 & 57.7  & 40.3 & 21.1 & 40.5 & 47.9 & \textbf{35.56} \\
  &\checkmark &\checkmark  &\checkmark  & 38.1 & 58.6  & 41.5 & 22.2 & 41.5 & 49.4 & \textbf{35.56}\\
 \checkmark&  &\checkmark  &\checkmark  & \textbf{38.6} & \textbf{60.1} & \textbf{42.0}  & \textbf{22.9} & \textbf{42.4} & 49.7 & 41.53\\
 \checkmark &\checkmark  &\checkmark  &\checkmark  & 38.4 & 59.7  & 41.9 & 22.3 & 42.0 & \textbf{49.8} & \textbf{35.56} \\

\hline
\end{tabular}
\end{center}}
\end{table*}

\begin{table*}[h]
\caption{Object detection results (bounding box AP) on COCO test-dev(\%)}
\label{tab:my-table10}
\setlength{\tabcolsep}{1.5mm}{
\begin{center}
\begin{tabular}{ccccccccc}
\hline
\multicolumn{1}{c}{\bf Method}  &\multicolumn{1}{c}{\bf Backbone}  &\multicolumn{1}{c}{\bf AP} &\multicolumn{1}{c}{\bf AP$_{50}$} &\multicolumn{1}{c}{\bf AP$_{75}$} &\multicolumn{1}{c}{\bf AP$_{s}$} &\multicolumn{1}{c}{\bf AP$_{m}$} &\multicolumn{1}{c}{\bf AP$_{l}$} &\multicolumn{1}{c}{\bf Params (M)} 
\\ \hline 
 Faster R-CNN& ResNet-50 & 37.7 & 58.7 & 40.8 &21.8 &40.6 &46.7 & 41.53   \\
 Libra R-CNN & ResNet-50&  38.6  & 60.0   & 42.0 & 22.4 & 41.3 & 47.7 & 41.79  \\
 Double Head & ResNet-50 & 39.8 & 60.2  & 43.4 & 23.0 & 42.7 & 49.8 & 47.12 \\
 Cascade R-CNN & ResNet-50 & 40.6 & 59.2 & 44.0  & 23.0 & 43.4 & 51.1 & 69.17 \\
 Faster R-CNN & ResNet-101& 39.7  & 60.7   & 43.3 & 22.6 & 42.9 & 49.9 & 60.52    \\
 Libra R-CNN & ResNet-101& 40.5  & 61.6   & 44.3 & \textbf{23.2} & 43.5 & 50.7 & 60.78 \\
 Double Head & ResNet-101 & 41.6 & 62.0 & 45.7 & 23.8 & 44.8 & 52.7 & 66.27\\
 Cascade R-CNN & ResNet-101 & 42.3 & 61.0  & 46.0 & 23.9 & 45.4 & 53.6 & 88.16 \\
  \hline
  Faster R-CNN(Ours)& ResNet-50 & 38.6$^\textbf{{+0.9}}$ & 60.3$^\textbf{{+1.6}}$ & 41.9$^\textbf{{+1.1}}$ &22.6$^\textbf{{+0.8}}$ &41.4$^\textbf{{+0.8}}$ &47.$^\textbf{{+1.2}}$ & 35.56$^\textbf{{-5.97}}$   \\
 Libra R-CNN(Ours) & ResNet-50& 39.1$^\textbf{{+0.5}}$  & 60.3$^\textbf{{+0.3}}$   & 42.8$^\textbf{{+0.8}}$ & 22.5$^\textbf{{+0.1}}$ & 41.6$^\textbf{{+0.3}}$ &48.5$^\textbf{{+0.8}}$ & 35.84$^\textbf{{-5.97}}$  \\
 Double Head(Ours) & ResNet-50 &40.3$^\textbf{{+0.5}}$ & 61.2$^\textbf{{+1.0}}$  &44.0$^\textbf{{+0.6}}$ &23.7$^\textbf{{+0.7}}$ &43.0$^\textbf{{+0.3}}$ &50.5$^\textbf{{+0.7}}$ &41.15$^\textbf{{-5.97}}$ \\
 Cascade R-CNN(Ours) & ResNet-50 &41.9$^\textbf{{+1.3}}$ &61.0$^\textbf{{+1.8}}$ &45.5$^\textbf{{+1.5}}$  &24.3$^\textbf{{+1.3}}$ &44.3$^\textbf{{+0.9}}$ &52.9$^\textbf{{+1.8}}$ &51.28$^\textbf{{-17.9}}$ \\

 Faster R-CNN(Ours) & ResNet-101& 40.7$^\textbf{{+1.0}}$  &62.3$^\textbf{{+1.6}}$   &44.4$^\textbf{{+1.1}}$ &23.3$^\textbf{{+0.7}}$ &43.9$^\textbf{{+1.0}}$ &51.3$^\textbf{{+1.4}}$ & 54.57$^\textbf{{-5.97}}$    \\
 Libra R-CNN(Ours) & ResNet-101 &40.6$^\textbf{{+0.1}}$ & 61.7$^\textbf{{+0.1}}$  & 44.5$^\textbf{{+0.2}}$ &22.7$^\textbf{{-0.5}}$ &43.5$^\textbf{{+0.0}}$ &51.1$^\textbf{{+0.4}}$ &54.83$^\textbf{{-5.97}}$  \\
 Double Head(Ours) & ResNet-101 &42.0$^\textbf{{+0.4}}$ &62.8$^\textbf{{+0.8}}$ &46.0$^\textbf{{+0.3}}$ &24.1$^\textbf{{+0.3}}$ &45.0$^\textbf{{+0.2}}$ &53.2$^\textbf{{+0.5}}$ &60.38$^\textbf{{-5.97}}$\\
 Cascade R-CNN(Ours) & ResNet-101 & 43.5$^\textbf{{+1.2}}$ &62.6$^\textbf{{+1.6}}$  &47.2$^\textbf{{+1.2}}$ &24.9$^\textbf{{+1.0}}$ & 46.2$^\textbf{{+0.8}}$ &55.3$^\textbf{{+1.7}}$ &70.27$^\textbf{{-17.9}}$ \\

\hline
\end{tabular}
\end{center}}
\end{table*}

{\bf Different combinations of AFE and CCR:} Both AFE and CCR we proposed are plug-and-play feature relationship enhancement modules, and their different arrangement order and different placement positions may have different effects on the performance of the model. With this in mind, we search different aggregation strategies for both modules with regard to the detection accuracy. In Table~\ref{tab:my-table6}, “FC2-AFE-CCR” indicates that the AFE and CCR modules are connected in series after the second FC layer, and the rest can be deduced by analogy. Note that in this experiment these two modules are only placed after the second shared FC layer in the RoI head. Analogously, “AFE-CCR-FC2” means that the AFE and CCR modules are connected in series before the second FC layer, and vice versa. Correspondingly, "AFE-FC2-CCR" means that the afe module is inserted before the FC2 layer, and the CCR module is inserted after it, and vice versa. 
Quite apart from that, we also explore two kinds of parallel structures of AFE and CCR modules. Analogously, "FC2-\{AFE,CCR\}" indicates that the AFE and CCR modules are connected in parallel after the FC2 layer, and the final output is the sum of the two. And "FC2-\{AFE\_cls,CCR\_reg\}" indicates that the output features of the AFE module are used for classification, and the output features of the CCR module are used for regression positioning, and vice versa. 
"AFE$\times2$-FC2-CCR$\times2$" means adding two AFE and CCR modules \emph{in series} before and after FC2, while "\{AFE,AFE\}-FC2-\{CCR,CCR\}" means adding two AFE and CCR modules \emph{in parallel} before and after FC2. Experimentally, we find that the "AFE-FC2-CCR" combination mode can obtain the best performance, so we use this manner as our default setting.

{\bf Effects of different spatial multipliers and channel multipliers: } As mentioned earlier, the spatial multiplier $\alpha$ and channel multiplier $\beta$ defined by us can flexibly adjust the parameters of the first FC layer in the two-stage method head network. And different $\alpha$ and $\beta$ values will bring different balances between precision and complexity, so we try different combinations of these two hyperparameters to explore this potential pattern, and the results are shown in Table~\ref{tab:my-table7}. \emph{In particular, we can reduce the number of parameters by 11.47M while remaining the comparable performance when we choose \{$\alpha$, $\beta$\}=\{2, None\}.} Foreshadowing our results, we adopt \{5, None\} in the following experiments for availability and productiveness.

{\bf Variant experiments for spatial reduction: } With the goal of illustrating the effectiveness of our proposed SR module, in Table~\ref{tab:my-table8}, we design three variant experiments for the spatial compression of RoI. The first uses a $3\times3$ convolutional layer for spatial reduction, denoted as "SR\_Conv3". The second uses a $3\times3$ group convolution with groups of 256, denoted as "SR\_Conv3\_Group". The third is that we directly set the output size of RoIAlign to $256\times5\times5$, denoted as "RoIAlign\_$5\times5$". Note that in this section, we use EGCA, AFE and CCR modules by default except for "RoIAlign\_$5\times5$". Consequently, the parameters of FC1 obtained from these three variant experiments are the same as when $\alpha=5$ in our method. But compared to our method, the first variant method takes into account all the surrounding adjacent features, while our method uses Conv1d and only considers the relationship between linear adjacent features. Compared to the first variant, the second variant method eliminates the influence of channel dimension correlation information. The third variant method does not add additional parameters, but it retains less original information. Our empirical studies show that the interaction with larger receptive field brings side effect on feature enhancement, and it is superfluous and unproductive to capture dependencies across all channels.

{\bf Effects of each component in our method: } In this section, we conduct ablative experiments to analyze the contribution of the proposed different components to the model performance. From Table~\ref{tab:my-table9}, we can see that except for the "SR" module, by assembling other modules we proposed can effectively improve the performance of the model. \emph{Notably, our model achieves the best result (+1.2\%) when using EGCA, AFE and CCR simultaneously, whereas the increase in parameters is almost negligible. Subsequently, after adding the "SR" module, our model parameters are significantly reduced by 5.97M, while the model accuracy is only reduced by 0.2\%. }

\begin{figure}[h]
\centering
\includegraphics[width=0.45\textwidth]{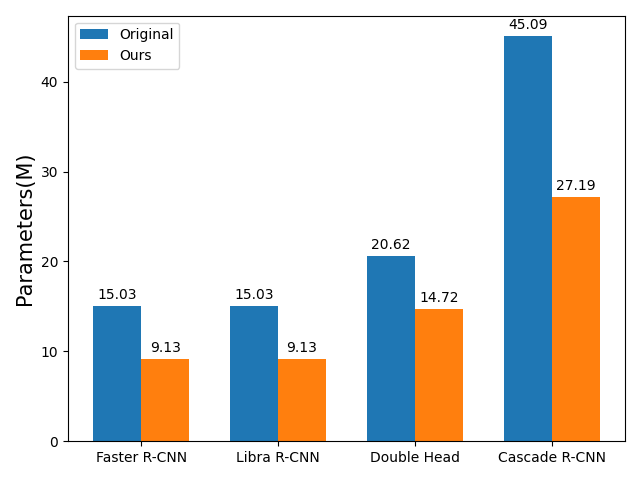}
   \caption{Comparison of head network parameters of different detectors. }
\label{fig:short}
\label{fig:fig6}
\end{figure}

\subsection{Main Results}
In this section, we compare our proposed method with state-of-the-art counterparts on COCO test-dev dataset. Without losing generality, we deploy our method on Cascade R-CNN, Libra R-CNN and Double Head R-CNN respectively to fully verify the effectiveness of our method. 

Without bells and whistles, Table~\ref{tab:my-table10} shows that our method can achieve continuous improvement on these four state-of-the-art object detection architectures with different backbone networks while significantly reducing the parameters of the detector head network (as shown in Fig.~\ref{fig:fig6}). Note that we use the AFE-FC2-AFE structure in the deployment of Cascade R-CNN, since the CCR module requires the number of RoIs and its feature dimensions to be consistent to meet the needs of the transposition operation, whereas the head network of Cascade R-CNN contains three stages and each stage has a different IoU threshold. Besides, our method is particularly prominent for the improvement of Cascade R-CNN, which can achieve a performance improvement of more than 1.2\% while reduce the number of parameters by 17.9M. And we conjecture that the reason is that our method can continuously drive the feature enhancement process at different stages in its head network to obtain better performance.

\section{Conclusion}
\label{sec:sec5}

In this paper, we propose CODH to learn the interactive relationship between local and global features of an instance, the autocorrelation of internal features of an instance, and the cross-correlation relationship between instances simultaneously. To this end, we use the lightweight EGCA module to supplement the global context information that is missing in the local receptive field of RoI. Simultaneously, we use the SR module to compress the features after RoIAlign cropping, which greatly reduces the complexity of the model. Moreover, we can model orthogonal autocorrelation and cross-correlation features by combining the proposed plug-and-play AFE and CCR modules. As a consequence, the performance of the detector can be effectively improved by combining the advantages of these different modules. 
The experimental results prove the efficiency and effectiveness of our method, and we also hope that our work can inspire other scholars.





\bibliographystyle{elsarticle-num} 
\bibliography{CODH}

\begin{thebibliography}{10}
\expandafter\ifx\csname url\endcsname\relax
  \def\url#1{\texttt{#1}}\fi
\expandafter\ifx\csname urlprefix\endcsname\relax\def\urlprefix{URL }\fi
\expandafter\ifx\csname href\endcsname\relax
  \def\href#1#2{#2} \def\path#1{#1}\fi

\bibitem{krizhevsky2017imagenet}
A.~Krizhevsky, I.~Sutskever, G.~E. Hinton, Imagenet classification with deep
  convolutional neural networks, Communications of the ACM 60~(6) (2017)
  84--90.

\bibitem{redmon2016you}
J.~Redmon, S.~Divvala, R.~Girshick, A.~Farhadi, You only look once: Unified,
  real-time object detection, in: Proceedings of the IEEE conference on
  computer vision and pattern recognition, 2016, pp. 779--788.

\bibitem{liu2016ssd}
W.~Liu, D.~Anguelov, D.~Erhan, C.~Szegedy, S.~Reed, C.-Y. Fu, A.~C. Berg, Ssd:
  Single shot multibox detector, in: European conference on computer vision,
  Springer, 2016, pp. 21--37.

\bibitem{lin2017feature}
T.-Y. Lin, P.~Doll{\'a}r, R.~Girshick, K.~He, B.~Hariharan, S.~Belongie,
  Feature pyramid networks for object detection, in: Proceedings of the IEEE
  conference on computer vision and pattern recognition, 2017, pp. 2117--2125.

\bibitem{sandler2018mobilenetv2}
M.~Sandler, A.~Howard, M.~Zhu, A.~Zhmoginov, L.-C. Chen, Mobilenetv2: Inverted
  residuals and linear bottlenecks, in: Proceedings of the IEEE conference on
  computer vision and pattern recognition, 2018, pp. 4510--4520.

\bibitem{ren2015faster}
S.~Ren, K.~He, R.~Girshick, J.~Sun, Faster r-cnn: Towards real-time object
  detection with region proposal networks, in: Advances in neural information
  processing systems, 2015, pp. 91--99.

\bibitem{pang2019libra}
J.~Pang, K.~Chen, J.~Shi, H.~Feng, W.~Ouyang, D.~Lin, Libra r-cnn: Towards
  balanced learning for object detection, in: Proceedings of the IEEE
  conference on computer vision and pattern recognition, 2019, pp. 821--830.

\bibitem{wu2020rethinking}
Y.~Wu, Y.~Chen, L.~Yuan, Z.~Liu, L.~Wang, H.~Li, Y.~Fu, Rethinking
  classification and localization for object detection, in: Proceedings of the
  IEEE/CVF conference on computer vision and pattern recognition, 2020, pp.
  10186--10195.

\bibitem{cai2018cascade}
Z.~Cai, N.~Vasconcelos, Cascade r-cnn: Delving into high quality object
  detection, in: Proceedings of the IEEE conference on computer vision and
  pattern recognition, 2018, pp. 6154--6162.

\bibitem{chen2019hybrid}
K.~Chen, J.~Pang, J.~Wang, Y.~Xiong, X.~Li, S.~Sun, W.~Feng, Z.~Liu, J.~Shi,
  W.~Ouyang, et~al., Hybrid task cascade for instance segmentation, in:
  Proceedings of the IEEE conference on computer vision and pattern
  recognition, 2019, pp. 4974--4983.

\bibitem{tian2019fcos}
Z.~Tian, C.~Shen, H.~Chen, T.~He, Fcos: Fully convolutional one-stage object
  detection, in: Proceedings of the IEEE international conference on computer
  vision, 2019, pp. 9627--9636.

\bibitem{he2017mask}
K.~He, G.~Gkioxari, P.~Doll{\'a}r, R.~Girshick, Mask r-cnn, in: Proceedings of
  the IEEE international conference on computer vision, 2017, pp. 2961--2969.

\bibitem{liu2018path}
S.~Liu, L.~Qi, H.~Qin, J.~Shi, J.~Jia, Path aggregation network for instance
  segmentation, in: Proceedings of the IEEE Conference on Computer Vision and
  Pattern Recognition, 2018, pp. 8759--8768.

\bibitem{huang2019mask}
Z.~Huang, L.~Huang, Y.~Gong, C.~Huang, X.~Wang, Mask scoring r-cnn, in:
  Proceedings of the IEEE Conference on Computer Vision and Pattern
  Recognition, 2019, pp. 6409--6418.

\bibitem{cheng2020boundary}
T.~Cheng, X.~Wang, L.~Huang, W.~Liu, Boundary-preserving mask r-cnn, in:
  European Conference on Computer Vision, Springer, 2020, pp. 660--676.

\bibitem{zhang2020mask}
W.~Zhang, C.~Fu, M.~Zhu, Mask point r-cnn, arXiv preprint arXiv:2008.00460
  (2020).

\bibitem{qiao2020detectors}
S.~Qiao, L.-C. Chen, A.~Yuille, Detectors: Detecting objects with recursive
  feature pyramid and switchable atrous convolution, arXiv preprint
  arXiv:2006.02334 (2020).

\bibitem{li2017light}
Z.~Li, C.~Peng, G.~Yu, X.~Zhang, Y.~Deng, J.~Sun, Light-head r-cnn: In defense
  of two-stage object detector, arXiv preprint arXiv:1711.07264 (2017).

\bibitem{hu2018relation}
H.~Hu, J.~Gu, Z.~Zhang, J.~Dai, Y.~Wei, Relation networks for object detection,
  in: Proceedings of the IEEE Conference on Computer Vision and Pattern
  Recognition, 2018, pp. 3588--3597.

\bibitem{wang2019region}
J.~Wang, K.~Chen, S.~Yang, C.~C. Loy, D.~Lin, Region proposal by guided
  anchoring, in: Proceedings of the IEEE Conference on Computer Vision and
  Pattern Recognition, 2019, pp. 2965--2974.

\bibitem{relationnetplusplus2020}
C.~Chi, F.~Wei, H.~Hu, Relationnet++: Bridging visual representations for
  object detection via transformer decoder, in: NeurIPS, 2020.

\bibitem{he2016deep}
K.~He, X.~Zhang, S.~Ren, J.~Sun, Deep residual learning for image recognition,
  in: Proceedings of the IEEE conference on computer vision and pattern
  recognition, 2016, pp. 770--778.

\bibitem{xie2017aggregated}
S.~Xie, R.~Girshick, P.~Doll{\'a}r, Z.~Tu, K.~He, Aggregated residual
  transformations for deep neural networks, in: Proceedings of the IEEE
  conference on computer vision and pattern recognition, 2017, pp. 1492--1500.

\bibitem{gao2019res2net}
S.~Gao, M.-M. Cheng, K.~Zhao, X.-Y. Zhang, M.-H. Yang, P.~H. Torr, Res2net: A
  new multi-scale backbone architecture, IEEE transactions on pattern analysis
  and machine intelligence (2019).

\bibitem{zhang2020resnest}
H.~Zhang, C.~Wu, Z.~Zhang, Y.~Zhu, Z.~Zhang, H.~Lin, Y.~Sun, T.~He, J.~Mueller,
  R.~Manmatha, et~al., Resnest: Split-attention networks, arXiv preprint
  arXiv:2004.08955 (2020).

\bibitem{howard2017mobilenets}
A.~G. Howard, M.~Zhu, B.~Chen, D.~Kalenichenko, W.~Wang, T.~Weyand,
  M.~Andreetto, H.~Adam, Mobilenets: Efficient convolutional neural networks
  for mobile vision applications, CoRR, abs/1704.04861 (2017).

\bibitem{zhang2018shufflenet}
X.~Zhang, X.~Zhou, M.~Lin, J.~Sun, Shufflenet: An extremely efficient
  convolutional neural network for mobile devices, in: Proceedings of the IEEE
  conference on computer vision and pattern recognition, 2018, pp. 6848--6856.

\bibitem{ma2018shufflenet}
N.~Ma, X.~Zhang, H.-T. Zheng, J.~Sun, Shufflenet v2: Practical guidelines for
  efficient cnn architecture design, in: Proceedings of the European conference
  on computer vision (ECCV), 2018, pp. 116--131.

\bibitem{tan2019efficientnet}
M.~Tan, Q.~Le,
  \href{http://proceedings.mlr.press/v97/tan19a.html}{{E}fficient{N}et:
  Rethinking model scaling for convolutional neural networks}, in:
  K.~Chaudhuri, R.~Salakhutdinov (Eds.), Proceedings of the 36th International
  Conference on Machine Learning, Vol.~97 of Proceedings of Machine Learning
  Research, PMLR, 2019, pp. 6105--6114.
\newline\urlprefix\url{http://proceedings.mlr.press/v97/tan19a.html}

\bibitem{han2020ghostnet}
K.~Han, Y.~Wang, Q.~Tian, J.~Guo, C.~Xu, C.~Xu, Ghostnet: More features from
  cheap operations, in: Proceedings of the IEEE/CVF Conference on Computer
  Vision and Pattern Recognition, 2020, pp. 1580--1589.

\bibitem{hu2018squeeze}
J.~Hu, L.~Shen, G.~Sun, Squeeze-and-excitation networks, in: Proceedings of the
  IEEE conference on computer vision and pattern recognition, 2018, pp.
  7132--7141.

\bibitem{li2019selective}
X.~Li, W.~Wang, X.~Hu, J.~Yang, Selective kernel networks, in: Proceedings of
  the IEEE conference on computer vision and pattern recognition, 2019, pp.
  510--519.

\bibitem{wang2020eca}
Q.~Wang, B.~Wu, P.~Zhu, P.~Li, W.~Zuo, Q.~Hu, Eca-net: Efficient channel
  attention for deep convolutional neural networks, in: Proceedings of the
  IEEE/CVF Conference on Computer Vision and Pattern Recognition, 2020, pp.
  11534--11542.

\bibitem{dai2017deformable}
J.~Dai, H.~Qi, Y.~Xiong, Y.~Li, G.~Zhang, H.~Hu, Y.~Wei, Deformable
  convolutional networks, in: Proceedings of the IEEE international conference
  on computer vision, 2017, pp. 764--773.

\bibitem{wang2018non}
X.~Wang, R.~Girshick, A.~Gupta, K.~He, Non-local neural networks, in:
  Proceedings of the IEEE conference on computer vision and pattern
  recognition, 2018, pp. 7794--7803.

\bibitem{zhu2019asymmetric}
Z.~Zhu, M.~Xu, S.~Bai, T.~Huang, X.~Bai, Asymmetric non-local neural networks
  for semantic segmentation, in: Proceedings of the IEEE International
  Conference on Computer Vision, 2019, pp. 593--602.

\bibitem{santoro2017simple}
A.~Santoro, D.~Raposo, D.~G. Barrett, M.~Malinowski, R.~Pascanu, P.~Battaglia,
  T.~Lillicrap, A simple neural network module for relational reasoning, in:
  Advances in neural information processing systems, 2017, pp. 4967--4976.

\bibitem{deng2019relation}
J.~Deng, Y.~Pan, T.~Yao, W.~Zhou, H.~Li, T.~Mei, Relation distillation networks
  for video object detection, in: Proceedings of the IEEE International
  Conference on Computer Vision, 2019, pp. 7023--7032.

\bibitem{kang2020graph}
J.~Kang, R.~Fernandez-Beltran, D.~Hong, J.~Chanussot, A.~Plaza, Graph relation
  network: Modeling relations between scenes for multilabel remote-sensing
  image classification and retrieval, IEEE Transactions on Geoscience and
  Remote Sensing (2020).

\bibitem{vaswani2017attention}
A.~Vaswani, N.~Shazeer, N.~Parmar, J.~Uszkoreit, L.~Jones, A.~N. Gomez,
  {\L}.~Kaiser, I.~Polosukhin, Attention is all you need, in: Advances in
  neural information processing systems, 2017, pp. 5998--6008.

\bibitem{zhang2020global}
W.~Zhang, C.~Fu, H.~Xie, M.~Zhu, M.~Tie, J.~Chen, Global context aware rcnn for
  object detection, arXiv preprint arXiv:2012.02637 (2020).

\bibitem{lin2014microsoft}
T.-Y. Lin, M.~Maire, S.~Belongie, J.~Hays, P.~Perona, D.~Ramanan,
  P.~Doll{\'a}r, C.~L. Zitnick, Microsoft coco: Common objects in context, in:
  European conference on computer vision, Springer, 2014, pp. 740--755.

\bibitem{mmdetection}
K.~Chen, J.~Wang, J.~Pang, Y.~Cao, Y.~Xiong, X.~Li, S.~Sun, W.~Feng, Z.~Liu,
  J.~Xu, Z.~Zhang, D.~Cheng, C.~Zhu, T.~Cheng, Q.~Zhao, B.~Li, X.~Lu, R.~Zhu,
  Y.~Wu, J.~Dai, J.~Wang, J.~Shi, W.~Ouyang, C.~C. Loy, D.~Lin, {MMDetection}:
  Open mmlab detection toolbox and benchmark, arXiv preprint arXiv:1906.07155
  (2019).

\end{thebibliography}
\biboptions{numbers,sort&compress}




\end{document}